\documentclass{article}
\usepackage{ijcai24}

\usepackage{times}
\usepackage{soul}
\usepackage{url}
\usepackage[hidelinks]{hyperref}
\usepackage[utf8]{inputenc}
\usepackage[small]{caption}
\usepackage{graphicx}
\usepackage{amsmath}
\usepackage{amsthm}
\usepackage{booktabs}
\usepackage{algorithm}
\usepackage{algorithmic}
\usepackage[switch]{lineno}
\usepackage{color}
\usepackage{enumitem}
\usepackage{amsfonts,amssymb} 
\usepackage{makecell}
\usepackage{multirow}
\usepackage{adjustbox}

\newcommand{\citet}[1]{\citeauthor{#1}~\shortcite{#1}}

\title{Understanding the planning of LLM agents: A survey}

\author{
Xu Huang$^1$
\and
Weiwen Liu$^2$\and
Xiaolong Chen$^1$\and
Xingmei Wang$^1$ \and
Hao Wang$^{1}$ \and \\
Defu Lian$^{1}$\footnote{Defu Lian is the corresponding author.} \and
Yasheng Wang$^2$ \and
Ruiming Tang$^2$ \and
Enhong Chen$^1$
\affiliations
$^1$University of Science and Technology of China, Hefei, China\\
$^2$Huawei Noah’s Ark Lab, Shenzhen, China
\emails
\{xuhuangcs, chenxiaolong, xingmeiwang\}@mail.ustc.edu.cn,
\{haowang, liandefu, cheneh\}@ustc.edu.cn,
\{liuweiwen8, wangyasheng, tangruiming\}@huawei.com
}
\begin{document}

\maketitle

\begin{abstract}
    As Large Language Models (LLMs) have shown significant intelligence, the progress to leverage LLMs as planning modules of autonomous agents has attracted more attention. This survey provides the first systematic view of LLM-based agents planning, covering recent works aiming to improve planning ability. We provide a taxonomy of existing works on LLM-Agent planning, which can be categorized into \textit{Task Decomposition}, \textit{Plan Selection}, \textit{External Module}, \textit{Reflection} and \textit{Memory}. Comprehensive analyses are conducted for each direction, and further challenges for the field of research are discussed.
\end{abstract}

\section{Introduction}

% [what is agent? what is planning ability in agent and why it matters?]
Autonomous agents have been recognized as intelligent entities capable of accomplishing specific tasks, via \textit{perceiving the environment}, \textit{planning}, and \textit{executing actions}.
Planning, as one of the most critical capabilities for agents, requires complicated understanding, reasoning, and decision-making progress~\cite{ghallab2004automated}.

% [some symbolic definition of planning]
Despite the abstract concept of planning, a general formulation of the planning tasks can be described as follows. Given time step $t$, with the environment denoted as $E$, the action space as $A$, the task goal as $g$, and the action at step $t$ as $a_{t}\in A$, the planning procedure can be expressed as the generation of a sequence of actions:
{\small\begin{align*}
    p = (a_0, a_1, \cdots, a_t) = \operatorname{plan}(E, g; \Theta, \mathcal{P}).
\end{align*}}
where $\Theta$ and $\mathcal{P}$ represent the parameters of the LLM and the prompts for the task, respectively.

% [from conventional methods to LLM-based methods]
Conventional works mainly rely on symbolic methods or reinforcement learning-based methods, such as Planning Domain Definition Language (PDDL)~\cite{aeronautiques1998pddl,haslum2019introduction} or policy learning~\cite{he2015deep,yao2020keep}. However, those conventional methods have several limitations. Symbolic methods require conversion from flexible natural language-described problems into symbolic modeling, which may require human experts' efforts. Usually, this kind of method lacks error tolerance, resulting in failures even if there are only a few errors. 
Reinforcement learning (RL) methods are often combined with deep models, which serve as the policy network or reward model. While RL algorithms often require a large number of samples (interactions with the environment) to learn an effective policy, this can be impractical or costly in scenarios where collecting data is time-consuming or expensive. 
% [weakness of conventional methods]

\begin{figure}[tb]
    \centering
    \includegraphics[width=0.75\columnwidth]{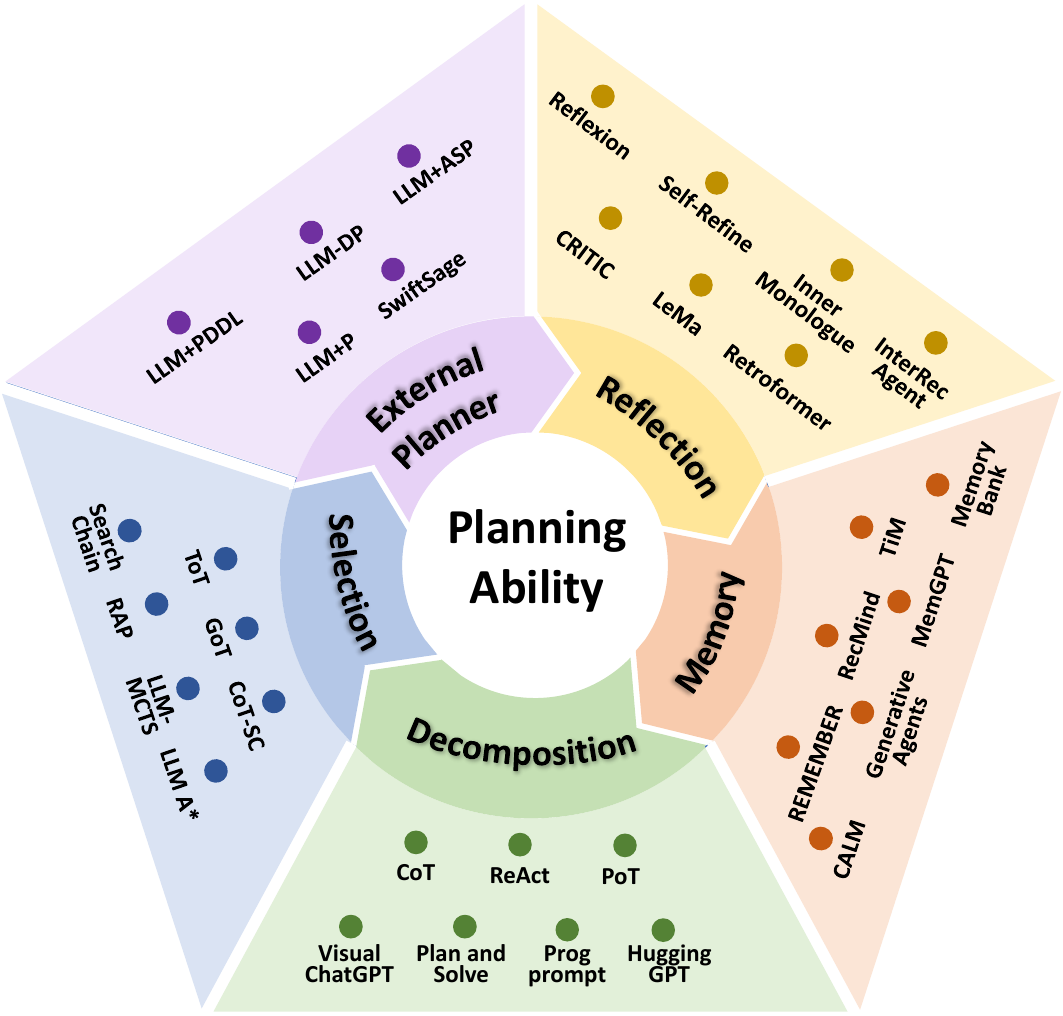}
    \caption{Taxonomy on LLM-Agent planning.}
    \label{fig:taxonomy}
\end{figure}

In recent years, the emergence of Large Language Models (LLMs) has marked a paradigm shift. LLMs have achieved remarkable success across various domains, showcasing significant intelligence in reasoning, tool usage, planning, and instruction-following. 
The surprising intelligence of LLMs sheds light on employing LLMs as the cognitive core of agents, thereby offering the potential to improve planning ability. Numerous methodologies have been developed to harness the potential of LLMs for agent planning. While existing surveys have attempted to summarize techniques for LLMs~\cite{zhao2023survey}, LLMs for decision-making~\cite{yang2023foundation}, reasoning~\cite{sun2023survey}, tool learning~\cite{qin2023tool}, and autonomous agents~\cite{wang2023survey}, they often lack a detailed analysis of planning ability within the literature.
In this survey, we analyze the latest research works and discuss the advantages and limitations, aiming to provide a systematic view of the planning ability of LLM-based agents. Existing methods are further categorized into five representative directions, with each direction undergoing comprehensive analysis. Furthermore, we have evaluated several representative methods on four benchmarks. To the best of our knowledge, this is the first work that comprehensively analyzes LLM-based agents from the planning abilities.

The subsequent sections of this paper are organized as follows. In Section~\ref{sec: tax}, we categorize the works into five mainstream directions and analyze their ideas regarding planning ability. Sections 3 to 7 provide detailed discussions and analysis of each direction. Finally, Section~\ref{sec: future} concludes the survey, offering insights into future directions in this field.

\section{Taxonomy}\label{sec: tax}
As the research on the planning ability of LLM-based agents presents a flourishing scene, various methods have been proposed to exploit the upper limit of planning ability. To have a better bird's view of existing advanced works, we pick out some representative and influential works, analyzing their motivations and essential ideas. To provide a better understanding, we illustrate the analysis in Table~\ref{tab:comp}. 

According to the table, we present a novel and systematic taxonomy for LLM-based agent plannning that divides existing works into five important categories, covering \textit{task decomposition}, \textit{multi-plan selection}, \textit{external module-aided planning}, \textit{reflection and refinement} and \textit{memory-augmented planning}, as illustrated in Figure~\ref{fig:taxonomy}. Here we briefly summarize those five directions as below.

% \begin{itemize}[leftmargin=*]
    \smallskip \noindent \textbf{Task Decomposition}. Tasks in real life are usually complicated and multi-step, bringing severe hardness for planning. This kind of method adopts the idea of \textit{divide and conquer}, decomposing the complicated into several sub-tasks and then sequentially planning for each sub-task. The process could be formulated as follows:
    {\small\begin{align*}
        g_0&, g_1, \cdots, g_n = \operatorname{decompose}(E, g;\Theta, \mathcal{P}); \\
        p^i &= (a^i_0,a^i_1,\cdots a^i_m) = \operatorname{sub-plan}(E, g_i;\Theta,\mathcal{P}).
    \end{align*}}
    
    \noindent \textbf{Multi-plan Selection}. This kind of method focuses on leading the LLM to ``think'' more, generating various alternative plans for a task. Then a task-related search algorithm is employed to select one plan to execute. The process could be formulated as follows:
    {\small\begin{align*}
        P &= p_1, p_2, \cdots, p_n = \operatorname{plan} (E, g;\Theta,\mathcal{P});\\
        p^* &= \operatorname{select}(E,g,P;\Theta,\mathcal{F}).
    \end{align*}}
    where $\mathcal{F}$ represents the search strategies, such as some tree search algorithms~\cite{yao2023ToT,zhao2023llmMCTS}.
    
    \smallskip\noindent \textbf{External Planner-Aided Planning}. This methodology is crafted to employ an external planner to elevate the planning procedure, aiming to address the issues of efficiency and infeasibility of generated plans, while the LLM mainly plays the role in formalizing the tasks. The process could be formulated as follows:
    {\small\begin{align*}
        h &= \operatorname{formalize}(E, g; \Theta, \mathcal{P}); \\
        p &= \operatorname{plan}(E,g,h;\Phi).
    \end{align*}}
    where $\Phi$ denotes the external planner module, $h$ represents the formalized information.
    
    \noindent \textbf{Reflection and Refinement}. This methodology emphasizes improving planning ability through reflection and refinement. It encourages LLM to reflect on failures and then refine the plan. The process could be formulated as follows:
    {\small\begin{align*}
        p_0 &= \operatorname{plan}(E,g;\Theta,\mathcal{P}); \\
        r_{i} &= \operatorname{reflect}(E,g,p_{i};\Theta,\mathcal{P}); \\
        p_{i+1} &= \operatorname{refine}(E,g,p_{i},r_{i};\Theta,\mathcal{P});
        % &\cdots \\
        % p_n &= \operatorname{refine}(E,g,p_{n-1},r_{n-1};\Theta,\mathcal{P}).
    \end{align*}}
    
    \smallskip \noindent \textbf{Memory-augmented Planning}. This kind of approach enhances planning with an extra memory module, in which valuable information is stored, such as commonsense knowledge, past experiences, domain-specific knowledge, et al. The information is retrieved when planning, serving as auxiliary signals. The process could be formulated as follows:
    {\small\begin{align*}
        m &= \operatorname{retrieve}(E, g;\mathcal{M}); \\
        p &= \operatorname{plan}(E,g,m;\Theta,\mathcal{P}).
    \end{align*}}
    where $\mathcal{M}$ represents the memory module.
% \end{itemize}

The five directions are interconnected rather than mutually exclusive, often involving the concurrent adoption of multiple techniques. In the subsequent sections, we delve deeper into the five research directions concerning LLM-agent planning, elucidating their motivations, proposing representation solutions, and addressing inherent limitations.

\begin{table*}[tbh]
    \caption{A taxonomy for existing LLM-Agent planning works.}
    % \scriptsize
    \centering
    \small
    \begin{adjustbox}{max width=0.95\linewidth}
    \setlength{\tabcolsep}{1.6mm}
    \begin{tabular}{c|c|c|c|c}
        \toprule
        Method & Idea & LLM's task & Formulation  & Representative works \\ \midrule
        \makecell[c]{Task\\Decomposition} & Divide and Conquer & \makecell[c]{Task decomposition\\Subtask planning} & \makecell[c]{$[g_i] = \operatorname{decompose}(E, g;\Theta, \mathcal{P});$ \\$p^i =  \operatorname{sub-plan}(E, g_i;\Theta,\mathcal{P})$} & \makecell[c]{CoT~\shortcite{wei2022CoT}, ReAct~\shortcite{yao2022react},\\HuggingGPT~\shortcite{shen2023hugginggpt}} \\ \midrule
        \makecell[c]{Multi-plan\\Selection} & \makecell[c]{Generate multiple plans \\and select the optimal} & \makecell[c]{Plans generation\\Plans evaluation} & \makecell[c]{$P = \operatorname{plan} (E, g;\Theta,\mathcal{P});$\\$p^* = \operatorname{select}(E,g,P;\Theta,\mathcal{F})$}  & \makecell[c]{ToT~\shortcite{yao2023ToT}, GoT~\shortcite{besta2023GoT},\\CoT-SC~\shortcite{wang2022selfconsistency}} \\ \midrule
        \makecell[c]{External\\Planner-aided} & \makecell[c]{Formalize tasks and\\ utilize external planner} & \makecell[c]{Task formalization} & \makecell[c]{$h = \operatorname{formalize}(E, g; \Theta, \mathcal{P});$ \\$p = \operatorname{plan}(E,g,h;\Phi)$}  & \makecell[c]{LLM+P~\shortcite{liu2023llm+},\\LLM+PDDL~\shortcite{guan2023leveraging}} \\ \midrule
        \makecell[c]{Reflection\\\& Refinement} & \makecell[c]{Reflect on experiences \\and refine plans} & \makecell[c]{Plan generation\\Reflection\\Refinement} & \makecell[c]{$p_0 = \operatorname{plan}(E,g;\Theta,\mathcal{P});$\\$r_{i} = \operatorname{reflect}(E,g,p_{i};\Theta,\mathcal{P});$ \\$p_{i+1} = \operatorname{refine}(E,g,p_{i},r_{i};\Theta,\mathcal{P})$}  & \makecell[c]{Reflexion~\shortcite{shinn2023reflexion},\\CRITIC~\shortcite{gou2023critic},\\Self-Refine~\shortcite{madaan2023selfrefine}} \\ \midrule
        \makecell[c]{Memory-aided\\Planning} & \makecell[c]{Leverage memory \\to aid planning} & \makecell[c]{Plan generation\\Memory extraction}  & \makecell[c]{$m = \operatorname{retrieve}(E, g;\mathcal{M}); $\\ $p = \operatorname{plan}(E,g,m;\Theta,\mathcal{P})$}  & \makecell[c]{REMEMBER~\shortcite{zhang2023remember},\\MemoryBank~\shortcite{zhong2023memorybank}}  \\
        \bottomrule
    \end{tabular}
    \end{adjustbox}
    \label{tab:comp}
\end{table*}

\section{Task Decomposition} \label{sec:decompose}
% [The basic idea of task decomposition]
% TODO: need to be simplified
In real-world scenarios, environments are often characterized by complexity and variability, thereby addressing complex tasks through a one-step planning process is a formidable challenge. 
% Take the organization of a large-scale academic conference, for instance, assigning such a complex task to an individual or an intelligent agent proves exceptionally difficult. This often results in various oversights in the planning process. In human society, these challenges are systematically decomposed into multiple sub-tasks for resolution. These sub-tasks may encompass issues such as venue selection, financial preparations, paper review processes, et al.

\begin{figure}[htb]
    \centering
    \includegraphics[width=0.7\columnwidth]{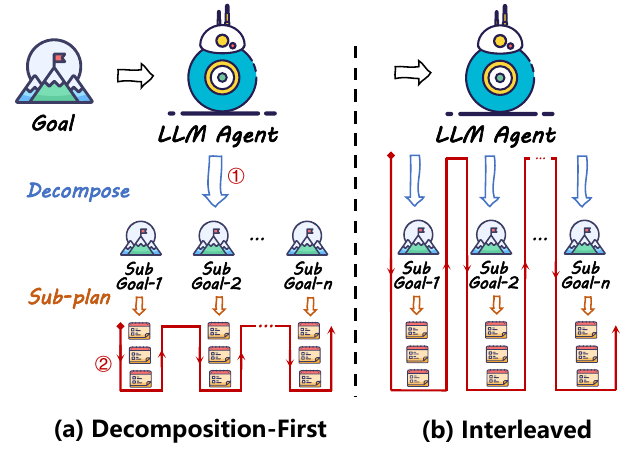}
    \caption{Types of task decomposition manners.}
    \label{fig:task-decomp}
\end{figure}

This simplification of complicated tasks is a remarkable human ability, evident in the decomposition of one task into several simpler sub-tasks~\cite{schraagen2000cognitive}, which is analogous to the well-known algorithmic strategy called ``divide and conquer", as illustrated in Eq. (1). Task decomposition generally involves two crucial steps: firstly, decomposing the complex task, referred to as the ``decompose" step, and secondly, planning for the sub-tasks, known as the ``sub-plan step".
 Current methods for task decomposition in this domain generally fall into two categories: decomposition-first and interleaved decomposition, illustrated in Figure~\ref{fig:task-decomp}.

\subsection{Decomposition-First Methods}
% [All related methods could be categorized into two types: decomposition first, interleaved decomposition and sub-planning]
% [First category: decomposition]
Decomposition-first methods decompose the task into sub-goals first and then plan for each sub-goal successively, presented in Figure~\ref{fig:task-decomp}(a). The representative methods include HuggingGPT~\cite{shen2023hugginggpt}, Plan-and-Solve~\cite{wang2023plansolve}, ProgPrompt~\cite{singh2023progprompt}, et al.

HuggingGPT~\cite{shen2023hugginggpt} utilizes various multimodal models from the Huggingface Hub to construct an intelligent agent for multimodal tasks. It is capable of handling tasks such as image generation, image classification, object recognition, video annotation, speech-to-text, et al. To facilitate collaboration between different models, the LLM acts as a controller, responsible for decomposing tasks inputted by humans, selecting models, and generating final responses. The most crucial stage is the initial task decomposition, where HuggingGPT explicitly instructs the LLM to break down the given task into sub-tasks, providing dependencies between tasks. 
Plan-and-Solve~\cite{wang2023plansolve} improves upon the Zero-shot Chain-of-Thought~\cite{kojima2022zeroCoT} by transforming the original ``\textit{Let's think step-by-step}" into a two-step prompt instruction: ``\textit{Let's first devise a plan}" and ``\textit{Let's carry out the plan}". This zero-shot approach has achieved improvements in mathematical reasoning, common-sense reasoning, and symbolic reasoning.
ProgPrompt~\cite{singh2023progprompt} translates natural language descriptions of tasks into coding problems. It symbolizes the agent's action space and objects in the environment through code, with each action formalized as a function and each object represented as a variable. Consequently, task planning is naturally transformed into function generation. When executing tasks, the agent first generates a plan in the form of function callings and then executes them step by step.

\subsection{Interleaved Decomposition Methods}
% [Second category: interleaved decomposition and sub-planning]
Interleaved decomposition involves interleaved task decomposition and sub-task planning, where each decomposition only reveals one or two sub-tasks at the current state, illustrated in Figure~\ref{fig:task-decomp}(b). Representative methods in this category include the Chain-of-Thought (CoT) series~\cite{wei2022CoT,kojima2022zeroCoT}, ReAct~\cite{yao2022react}, PAL~\cite{gao2023pal}, Program-of-Thought (PoT)~\cite{chen2022program}, Visual ChatGPT~\cite{wu2023visualchatgpt}, et al.

The introduction of Chain-of-Thought (CoT)~\cite{wei2022CoT} reveals the few-shot learning capabilities of LLM. CoT guides the LLM in reasoning about complex problems through a few constructed trajectories, leveraging the LLM's reasoning abilities for task decomposition. 
% In a step-by-step manner, LLM conducts task decomposition, providing the next sub-task goal after each step and waiting for the agent to complete it. It alternates between the decompose and sub-plan steps.
Subsequently, Zero-shot CoT~\cite{kojima2022zeroCoT} unlocks the LLM's zero-shot reasoning abilities with the magical instruction ``\textit{Let's think step-by-step}".
In contrast to CoT, which embeds reasoning within the planning process, ReAct~\cite{yao2022react} decouples reasoning and planning. It alternates between reasoning (Thought step) and planning (Action step), demonstrating significant improvements in the planning capabilities.
Visual ChatGPT~\cite{wu2023visualchatgpt} utilizes ReAct's mechanism, employing LLM as the agent's brain equipped with a series of visual models, resulting in an agent with image processing capabilities.
PAL~\cite{gao2023pal} improves CoT by leveraging the LLM's coding abilities, guiding the LLM to generate code during reasoning. Finally, a code interpreter (such as Python) is used to comprehensively execute the codes to obtain the solution. This method proves helpful for agents in solving mathematical and symbolic reasoning problems.
Program-of-Thought (PoT)~\cite{chen2022program} completely formalize the reasoning process as programming. The authors also leverage a CodeX~\cite{chen2021codex} model trained on code-related data, enhancing performance in mathematical and financial problems.

% [Discussion about the advantages and limitations]
\subsection{Discussions} 
% Regarding the two categories of task decomposition, we have summarized their advantages and limitations, respectively. 
For the decomposition-first method, the advantage lies in creating a stronger correlation between the sub-tasks and the original tasks, reducing the risk of task forgetting and hallucinations~\cite{touvron2023llama}. However, since the sub-tasks are predetermined at the beginning, additional mechanisms for adjustment are required otherwise one error in some step will result in failure, which will be discussed in Section~\ref{sec:refine}.
On the other hand, interleaved decomposition and sub-planning dynamically adjust decomposition based on environmental feedback, improving the fault tolerance. However, for complicated tasks, excessively long trajectories may lead to LLM experiencing hallucinations, deviating from the original goals during subsequent sub-tasks and sub-planning.

% [Challenges in task decomposition]
Although task decomposition significantly enhances the ability of LLM-Agent to solve complicated tasks, challenges persist. The first challenge is the additional overhead introduced by task decomposition. Decomposing a task into multiple sub-tasks requires more reasoning and generation, incurring additional time and computational costs. On the other hand, for highly complex tasks that are decomposed into dozens of sub-tasks, the planning is constrained by the context length of the LLM, leading to the forgetting of the planning trajectories. 
% Additionally, excessively long trajectories can lead to LLM experiencing task forgetting and hallucinations.
\section{Multi-Plan Selection} \label{sec:select}
Due to the complexity of the tasks and the inherent uncertainty of LLM, the plans generated by the LLM-Agent for a given task can be diverse. Even though LLM possesses strong reasoning abilities, a single plan generated by LLM is likely to be suboptimal or even infeasible. A more natural approach is multi-plan selection, comprising two major steps: multi-plan generation and optimal plan selection.

\subsection{Multi-Plan Generation}
Multi-plan generation involves generating a dozen paths of plans to comprise the candidate plan set. Mainstream methods consider employing uncertainty in the decoding process of generative models.

Self-consistency~\cite{wang2022selfconsistency} employs a simple intuition: the solutions for complex problems are rarely unique. In contrast to CoT, which generates a single path, Self-consistency obtains multiple distinct reasoning paths via sampling strategies embodied in the decoding process, such as temperature sampling, top-\textit{k} sampling. 
Tree-of-Thought (ToT)~\cite{yao2023ToT} proposes two strategies to generate plans (i.e. thoughts): sample and propose. The \textit{sample} strategy is consistent with Self-consistency, where LLM would sample multiple plans in decoding process. The \textit{propose} strategy explicitly instructs the LLM to generate various plans via few-shot examples in prompts. 
Graph-of-Thought (GoT)~\cite{besta2023GoT} extends ToT by adding transformations of thoughts, which supports arbitrary thoughts aggregation. 
LLM-MCTS~\cite{zhao2023llmMCTS} and RAP~\cite{hao2023rap} leverages LLM as the heuristic policy function for the Monte Carlo Tree Search (MCTS), where multiple potential actions are obtained by multiple calls. 

\subsection{Optimal Plan Selection}
To select the optimal plan among the candidate plans, diverse strategies are adopted as heuristic search algorithms. 

Self-consistency~\cite{wang2022selfconsistency} applies the naive majority vote strategy, regarding the plan with the most votes as the optimal choice. Benefiting from the tree architecture, Tree-of-Thought (ToT)~\cite{yao2023ToT} supports tree search algorithms, such as conventional BFS and DFS. When selecting a node for expansion, it uses LLM to evaluate multiple actions and chooses the optimal one. Similar with ToT, LLM-MCTS~\cite{zhao2023llmMCTS} and RAP~\cite{hao2023rap} also employ a tree structure to assist in multi-plan search. Unlike ToT, they employ the Monte Carlo Tree Search (MCTS) algorithm for search. LLM A*~\cite{xiao2023llmAstar} utilizes the classic A* algorithm from artificial intelligence to assist LLM in search. The Chebyshev distance from the current position to the target position serves as the heuristic cost function for selecting the optimal path.

\subsection{Discussions} 
The scalability of multi-plan selection is notably advantageous, providing a broader exploration of potential solutions in the expansive search space. However, this advantage comes with inherent trade-offs. The increased computational demands, especially for models with large token counts or computations, pose practical challenges. This cost consideration becomes crucial, particularly in scenarios where resource constraints are a significant factor, such as the online service.
Moreover, the reliance on LLM for the evaluation of plans introduces new challenges. As LLM's performance in ranking tasks is still under scrutiny, there is a need for further validation and fine-tuning of its capabilities in this specific context. The stochastic nature of LLMs adds randomness to the selection, potentially affecting the consistency and reliability of the chosen plans.
\section{External Planner-Aided Planning}    \label{sec:module}
Despite the powerful reasoning and task decomposition capabilities exhibited by Large Language Models (LLMs), challenges arise when confronted with environments featuring intricate constraints, such as mathematical problem-solving or generating admissible actions. 
% The conventional prompt techniques often fall short in generating complete descriptions for those constraints, rendering LLMs challenged in making plans that adhere to such constraints. For instance, in the context of embodied agents, LLMs may produce actions that are presently impractical within the given environment.
To address challenges, several methods integrate LLMs with external planners. Such methods can be categorized into symbolic planners and neural planners based on the introduced planners.

\subsection{Symbolic Planner}
Symbolic planners have served as a fundamental component in the fields of automated planning for several decades. These approaches, based on well-established symbolic formalized models, such as PDDL models~\cite{aeronautiques1998pddl,haslum2019introduction}, employ symbolic reasoning to identify optimal paths from initial states to desired goal states.

LLM+P~\cite{liu2023llm+} enhances the planning proficiency of LLMs by incorporating a PDDL-based symbolic planner. Leveraging the semantic understanding and coding capabilities of LLM, the authors organize problems into textual language prompts inputted to LLM. This prompts LLM to organize the actions within the environment and specified tasks into the format of the PDDL language. Subsequently, after obtaining a formalized description, the authors employ the Fast-Downward~\footnote{\url{https://github.com/aibasel/downward/tree/release-22.12.0}} solver for the planning process.
Building upon LLM+P, LLM-DP~\cite{dagan2023dynamic} is specifically designed for dynamic interactive environments. Upon receiving feedback from the environment, LLM processes the information, formalizes it into PDDL language, and then employs a BFS~\cite{lipovetzky2014width} solver to generate a plan.
LLM+PDDL~\cite{guan2023leveraging} also utilizes the PDDL language to formalize the task, incorporating an additional step for manual verification to check for potential issues in the PDDL model generated by LLM. During the planning process, the authors propose using the plan generated by LLM as an initial heuristic solution to accelerate the search process of local search planners, such as LPG~\cite{gerevini2002lpg}.
LLM+ASP~\cite{yang2023coupling} transforms problems described in natural language by LLM into atomic facts, converting tasks into ASP problems. Subsequently, the ASP solver CLINGO is utilized to generate plans.

\subsection{Neural Planner}
Neural planners are deep models trained on collected planning data with reinforcement learning or imitation learning techniques, showing effective planning abilities within the specific domain. For instance, DRRN~\cite{he2015deep} models the planning process as a Markov Decision Process through reinforcement learning, training a policy network to obtain a deep decision model. Decision Transformer (DT)~\cite{chen2021decision} empowers a transformer model to clone human decision-making behavior with planning data.

Well-trained neural planners exhibit excellent planning capabilities within their respective domains and demonstrate superior planning efficiency due to their smaller parameter sizes. However, when faced with complex and less frequently encountered problems, where training data is scarce, these small models tend to perform poorly due to insufficient generalization ability. Therefore, several works explore combining an LLM with a light-weight neural planner, to further enhance the planning capabilities. CALM~\cite{yao2020keep} proposed an early approach that combines a language model with an RL-based neural planner. One language model processes textual environmental information, generating a set of candidate actions as priors based on the environmental information. A DRRN policy network is then employed to re-rank these candidate actions, ultimately selecting the optimal action. SwiftSage~\cite{lin2023swiftsage} leverages the dual-process theory from cognitive psychology, dividing the planning process into slow thinking and fast thinking. The slow-thinking process involves complex reasoning and rational deliberation while fast-thinking resembles an instinctive response developed through long-term training. The authors utilize a DT model, trained through imitation learning, as the fast-thinking model for rapid plan generation. When errors occur during plan execution, indicating a more complex problem, the agent switches to the slow-thinking process, where LLM engages in reasoning and planning based on the current state. This combination of fast and slow thinking has proven to be highly effective in terms of efficiency.

\subsection{Discussions}
For those strategies that leverage an additional planner for assistance, LLM primarily plays a supportive role. Its main functions involve parsing textual feedback and providing additional reasoning information to assist in planning, particularly when addressing complex problems. Specifically, the enhancement of LLM's capabilities in code generation empowers the potential to deal with more general tasks for symbolic artificial intelligence. Actually, a significant drawback of traditional symbolic AI systems lies in the complexity and heavy reliance on human experts in constructing symbolic models, while LLM accelerates this process, facilitating faster and more optimal establishment of symbolic models. The advantages brought by symbolic systems include theoretical completeness, stability, and interpretability. The combination of statistical AI with LLM is poised to become a major trend in the future development of artificial intelligence.

\section{Reflection and Refinement} \label{sec:refine}

Reflection and refinement are indispensable components in the planning process. 
They enhance the fault tolerance and error correction capabilities of LLM-Agent planning. Due to existing hallucination issues and insufficient reasoning abilities for complex problems, LLM-Agents may make errors and get stuck in ``thought loops'' during planning due to limited feedback. Reflecting on and summarizing failures helps agents correct errors and break out of such loops in subsequent attempts.

% \subsection{Reflection}

Self-refine~\cite{madaan2023selfrefine} utilizes an iterative process of generation, feedback, and refinement. After each generation, LLM generates feedback for the plan, facilitating adjustments based on the feedback.
Reflexion~\cite{shinn2023reflexion} extends ReAct by incorporating an evaluator to assess trajectories. LLM generates self-reflections upon error detection, aiding in error correction.
CRITIC~\cite{gou2023critic} uses external tools like Knowledge Bases and Search Engines to validate LLM-generated actions. It then leverages external knowledge for self-correction, significantly reducing factual errors.
InteRecAgent~\cite{huang2023interecagent} employs a mechanism called ReChain for self-correction. An LLM is used to evaluate the response and tool-using plan generated by the interactive recommendation agent, summarize feedback on errors, and decide whether to restart planning.
LEMA~\cite{an2023lema} gathers mistaken planning samples first and employs more powerful GPT-4 for correction. Those corrected samples are then used to fine-tune the LLM-Agent, resulting in significant performance improvements across various scales of the LLaMA model.

Particularly, the self-reflective strategy bears resemblance to the principles of reinforcement learning, where the agent plays the role of the decision-maker, such as the policy network. Environmental feedback triggers updates of the policy network. However, in contrast to deep reinforcement learning where updates are achieved by modifying model parameters, in the LLM agent, this update occurs through self-reflection by the LLM itself, culminating in textual verbal feedbacks. These textual feedbacks  can serve as both long-term and short-term memory, influencing the agent's subsequent planning outputs through the prompts. Nevertheless, the convergence of this textual form of update currently lacks a guaranteed proof, indicating the inability to demonstrate that continual reflection can ultimately lead the LLM agent to a specified goal.

\section{Memory-Augumented Planning} \label{sec:memory}

% One of the distinct characteristics of humanity is its capacity for growth: the ability to improve oneself from past experiences. This growth benefits a lot from human memory. 
For agents, memory is a crucial pathway to enhance planning capabilities and the potential for growth.
Regarding the memory mechanisms in LLM-Agents, there are currently two major approaches to enhance planning abilities through memory: RAG-based memory and embodied memory.

\subsection{RAG-based Memory}
Retrieval Augmented Generation (RAG)~\cite{lewis2020RAG,mao2020RAG4QA,cai2022RAGSurvey} techniques are proposed to aid text generation with retrieved information. It is capable of enhancing the LLM with the latest knowledge, such as New Bing\footnote{https://www.bing.com/} and Google Bard\footnote{https://bard.google.com}. For LLM agents, past experiences could be stored in the memory and retrieved when needed. The core idea of such methods is to retrieve task-relevant experiences from the memory during task planning. Among those methods, memories are typically stored in additional storage, and the forms are diverse, such as texts~\cite{park2023generativeagents,liu2023TiM,packer2023memgpt,wang2023recmind,zhong2023memorybank}, tabular forms~\cite{zhang2023remember}, knowledge graph~\cite{pan2024unifying}, etc.

Generative Agents~\cite{park2023generativeagents} store the daily experiences of human-like agents in text form and retrieve memories based on a composite score of recency and relevance to the current situation. Similarly, MemoryBank~\cite{zhong2023memorybank}, TiM~\cite{liu2023TiM}, and RecMind~\cite{wang2023recmind} encode each memory using a text encoding model into a vector and establish an indexing structure, such as FAISS library~\cite{johnson2019faiss}. During retrieval, the description of the current status is used as a query to retrieve memories from the memory pool. The difference between the three lies in the way memories are updated. MemGPT~\cite{packer2023memgpt} leverages the concept of multiple levels of storage in computer architecture, abstracting the context of LLM into RAM and treating the additional storage structure as a disk. LLM can spontaneously decide whether to retrieve historical memories or save the current context to storage. REMEMBER~\cite{zhang2023remember} stores historical memories in the form of a Q-value table, where each record is (environment, task, action, Q-value)-tuple. During retrieval, positive and negative memories are both retrieved for LLM to generate plan based on the similarity of the environment and task.

\subsection{Embodied Memory}
Embodied memory involves finetuning the LLM with the agent's historical experiential samples, embedding memories into the model parameters. Usually the experiential samples are collected from the agents's interactions with environment, which may consist of commonsense knowledge about the environment, task-related priors, and successful or failed experiences. While the cost of training a language model with more than billions of parameters is huge, parameter-efficient fine-tuning (PEFT) techniques are leveraged to reduce cost and speed up by training a small part of parameters only, such as LoRA, QLoRA, P-tuning, et al.

CALM~\cite{yao2020calm} utilizes ground-truth action trajectories collected from the text-world environment to finetune GPT-2 using next token prediction task, enabling it to memorize planning-related information and generalize well on planning tasks. Similarly, TDT~\cite{wang2022scienceworld} uses collected Markov decision process data to fine-tune Text Decision Transformer (TDT). It achieves better success rates on more challenging ScienceWorld~\cite{wang2022scienceworld} tasks. AgentTuning~\cite{zeng2023agenttuning} organizes plan trajectories from various tasks into a dialogue form to finetune the LLaMA model, showing significant improvements in performance on unseen planning tasks.

\subsection{Discussions} 
The RAG-based and Fine-tuning-based memory approaches enhance LLM-Agent planning capabilities, each with distinct advantages and limitations. RAG-based methods offer real-time, low-cost external memory updates mainly in natural language text, but rely on the accuracy of retrieval algorithm. Finetuning provides a larger memorization capacity through parameter modifications but has high memory update costs and struggles with retaining fine-grained details.

Memory-enhanced LLM-Agents demonstrate enhanced growth and fault tolerance in planning, yet memory generation heavily depends on LLM's generation capabilities. Improving weaker LLM-Agents through self-generated memory remains a challenging area to explore.
\section{Evaluation} \label{sec: eval}
% [representative benchmarks]: 
% type1: text-world interactive environment: AlfWorld, TWC, ScienceWorld,
% type2: Openworld QA environment: HotpotQA, FEVER
% type3: visual openworld games: MineCraft, 
% type4: Web related: WebShop, Mind2Web, WebArena\cite{zhou2023webarena}
% type5: computer skills: Operating System(AgentBench); Database(AgentBench); MiniWoB++\cite{kim2023miniwob};
% 

\begin{table}[t]
    \caption{Evaluation of representative prompt-based methods on four interactive benchmarks. The SR, AR and EX are abbreviations of success rate, average rewards, and expenses respectively. The expenses are calculated based on the number of consumed tokens through OpenAI's API. Z-CoT and F-CoT represent Zeroshot-CoT and Fewshot-CoT, respectively. }
    \centering
    \begin{adjustbox}{max width=\linewidth}
    \begin{tabular}{l|rr|rr|rr|rr}
        \toprule
            & \multicolumn{2}{c|}{AlfWorld} & \multicolumn{2}{c|}{ScienceWorld} & \multicolumn{2}{c|}{HotPotQA} & \multicolumn{2}{c}{FEVER}  \\ \midrule
         Metrics & SR(\%) & EX(\$) & AR & EX(\$) & SR(\%) & EX(\$) & SR(\%) & EX(\$)  \\ \midrule
         Z-CoT & N/A & N/A & N/A & N/A & 0.01 & 0.95 & 0.39 & 1.07  \\
         F-CoT & 0.43 & 98.60 & 16.58 & 272.22 & 0.32 & 5.73 & 0.61 & 2.25 \\
         CoT-SC & 0.57 & 105.37 & 15.24 & 274.33 & 0.33 & 7.86 & 0.62 & 3.21\\
         SayCan & 0.60 & 113.61 & 12.36 & 125.71 & N/A & N/A & N/A & N/A \\
         ReAct & 0.57 & 152.18 & 15.05 & 356.03 & 0.34 & 66.00 & 0.63 & 22.20 \\
         Reflexion & 0.71 & 220.17 & 19.39 & 724.48 & 0.39 & 112.49 & 0.68 & 37.26\\
         % PAL & & \\
         \bottomrule
    \end{tabular}
    \end{adjustbox}
    \label{tab: exp}
\end{table}

% Evaluation plays a crucial role in comparison between various methods. We 
Evaluating the planning capability of the agent is a critical issue in the research area. Here we investigate several mainstream benchmarking methods, categorizing them into the following types.

% \begin{itemize}[leftmargin=*]
    \smallskip \noindent \textit{Interactive Gaming Environments}: Game environments may provide real-time multi-modal feedback based on the agent's actions, including textual and visual feedback. Currently, the most widely used gaming environment is Minecraft~\footnote{\url{https://www.minecraft.net/}}, where the agent needs to gather materials to create tools for obtaining more rewards. The quantity of tools created by the agent is often used as an evaluation metric. Another popular category is the text-based interactive environments, such as ALFWorld~\cite{shridhar2020alfworld}, ScienceWorld~\cite{wang2022scienceworld}, et al, where the agent locates in an environment described in natural language, with limited actions and locations. The success rate or the rewards obtained are commonly used as evaluation metrics. Compared with Minecraft, these text-based interactive environments are often simpler, with straightforward feedback and fewer feasible actions.
    
    \smallskip \noindent \textit{Interactive Retrieval Environments}: Interactive retrieval environments simulate the process of information retrieval and reasoning that humans undergo in real life. In these environments, agents are often allowed to interact with search engines and other web services, using actions such as searching keywords or executing click, forward, and backward operations to acquire more information, thereby obtaining answers to questions or completing information retrieval tasks. Commonly used retrieval environments include question-answering tasks based on the Wikipedia engine~\cite{yao2022react} (such as HotPotQA~\cite{yang2018hotpotqa} and Fever~\cite{thorne2018fever}) and web browsing tasks to find specific information, including WebShop, Mind2Web~\cite{deng2023mind2web}, and WebArena~\cite{zhou2023webarena}. The task success rate is usually used as the metric.
    
    \smallskip \noindent \textit{Interactive Programming Environments}: Interactive programming environments simulate the interaction between programmers and computers, testing the agent's planning ability in solving computer-related problems. In these environments, agents are required to interact with computers to solve problems by writing code or instructions. They would receive various feedback including compile and runtime error messages, as well as execution results. Popular interactive programming environments involve issues related to operating systems, databases, etc., such as Agent Bench~\cite{liu2023agentbench}, MiniWoB++~\cite{kim2023miniwob}.
% \end{itemize}

Most of these existing interactive environments lack fine-grained evaluation, where the performance is predominantly evaluated by the final success rate. Furthermore, unlike real-world scenarios where there are often multiple paths to complete a task, there is typically only one ``golden'' path in most simulated environments due to the high annotation cost.

\smallskip
\noindent
\textbf{Experiments.} We have conducted experiments on four benchmarks to validate the performance of representative works, shown in Table~\ref{tab: exp}. We have implemented six prompt-based methods due to limited budgets, covering task decomposition, multi-path selection, and reflection. As for the benchmarks, ALFWorld, ScienceWorld, HotPotQA, and FEVER are employed, involving interactive gaming and question-answering benchmarks. 
Since ALFWorld and ScienceWorld are involved in larger action space, the zero-shot method, i.e. ZeroShot-CoT, is not applicable due to unawareness of action space. SayCan improves CoT by grounding output actions into action space with a value function, which does not apply to QA tasks because there are only two actions: \textsc{Search[keyword]} and \textsc{Lookup[keyword]}. And we set the value function as a textual embedding model \textsl{bge-small-en-v1.5}~\cite{bge_embedding}. 
We obtain 3 actions and 5 answers each step for gaming tasks and QA tasks for CoT-SC, respectively. The round of retries in Reflexion is set to 1. We use the API of \textsl{text-davinci-003} in OpenAI as LLM.
% We evaluate those methods with two metrics: success rate (or rewards) for planning effectiveness and expenses of consumed tokens. Based on the results, we have the following observations.
% We sample about 30 tasks from boil, mix, find and freeze tasks in ScienceWorld, resulting in 296 tasks in total. As for HotPotQA and FEVER, we sample 500 tasks from development set referring to ReAct~\cite{yao2022react}. S

\smallskip
\noindent
\textit{(i) The performance increases with the expenses.} As CoT-SC, ReAct and Reflexion are involved in multiple plans, additional thoughts, and reflections, respectively, their expenses are more than their backbone methods. Intuitively, more tokens represent more detailed thinking, resulting in performance improvements.

\smallskip
\noindent
\textit{(ii) Fewshot examples are suggested for complicated tasks.} Despite that the magic instruction \textit{Let's think step by step} can lead to more reasoning, ZeroShot-CoT exhibits severe performance degradation in two QA benchmarks, which demonstrates the necessity of the examples for LLM to further understand the task.

\smallskip
\noindent
\textit{(iii) Reflection plays a crucial role in improving the success rate, especially for complex tasks.} Despite Reflexion consuming about twice the tokens compared with ReAct, the improvements in complicated tasks are promising, such as ALFWorld and ScienceWorld, which shows that LLM possesses the error-correcting capability. 
% Furthermore, the reflection summarized from failures serves as a long-term memory in the next try, enabling the agent to learn from its failures.

\section{Conclusions and Future Directions} \label{sec: future}
Since LLM has shown the emergence of intelligence, there has been an increasing focus on using LLM to enhance the planning capabilities of agents.
The major directions are summarized in Figure 1, with a detailed comparison and analysis of various methods presented in Sections 3 to 7.
We also conducted experiments on four benchmarks, comparing the effectiveness of several representative methods and showing that performance increases with expenses. Despite the enhancements made by these works in planning capabilities, there are still some significant challenges.

\smallskip
\noindent
\textbf{Hallucinations.} During the planning process, LLM often suffers from hallucinations, leading to irrational plans, unfaithfulness to task prompts, or failing to follow complex instructions. For instance, plans may include actions that interact with items not existed in the environment. Although these issues can be alleviated through careful prompt engineering, they reflect fundamental shortcomings in LLM~\cite{zhang2023siren,huang2023survey}.

\smallskip
\noindent
\textbf{Feasibility of Generated Plans.} LLM, being fundamentally based on statistical learning, optimizes the probability of the next word through massive data. Compared to symbolic artificial intelligence, this approach struggles to obey complex constraints, especially when dealing with less common constraints encountered during LLM training. Consequently, plans generated by LLM may lack feasibility without considering adequate preconditions. Connecting LLM with symbolic planning models without altering LLM itself is a promising future direction.

\smallskip
\noindent
\textbf{Efficiency of Generated Plans.} Generating efficient plans is a crucial issue in planning. However, in existing LLM agents, planning is greedily based on generated plans from LLM output, without considering the efficiency of the generated plans. Therefore, future developments may require introducing additional efficiency evaluation modules to work in conjunction with LLM for more efficient plans.

\smallskip
\noindent
\textbf{Multi-Modal Environment Feedback.} LLM is originally designed for processing textual inputs, but real-world environment feedback is often multi-modal, including images, audio, etc., which are challenging to describe in natural language. Therefore, LLM agents face limitations when handling such scenarios. Future considerations may involve integrating the development of multi-modal large models and revisiting related planning strategies.

\smallskip
\noindent
\textbf{Fine-grained Evaluation.} As mentioned in Section~\ref{sec: eval}, existing benchmarks mostly rely on the final completion status of tasks, lacking fine-grained step-wise evaluations. Additionally, environmental feedback is often rule-based, simplistic, and distant from real-world scenarios. A potential future direction is to leverage high-intelligence models like LLM to design more realistic evaluation environments.

\bibliographystyle{named}
\bibliography{agent}

\end{document}